# Acronym Recognition and Processing in 22 Languages

**Maud Ehrmann**
Department of Computer Science
Sapienza University of Rome
`ehrmann@di.uniroma1.it`

**Leonida Della Rocca**
European Commission
Joint Research Centre
IPSC-GlobeSec
`leonida.della-rocca@ext.jrc.ec.europa.eu`

**Ralf Steinberger**
European Commission
Joint Research Centre
IPSC-GlobeSec
`ralf.steinberger@jrc.ec.europa.eu`

**Hristo Tannev**
European Commission
Joint Research Centre
IPSC-GlobeSec
`hristo.tannev@jrc.ec.europa.eu`

## Abstract

We are presenting work on recognising acronyms of the form *Long-Form (Short-Form)* such as "*International Monetary Fund (IMF)*" in millions of news articles in twenty-two languages, as part of our more general effort to recognise *entities* and their variants in news text and to use them for the automatic analysis of the news, including the linking of related news across languages. We show how the acronym recognition patterns, initially developed for medical terms, needed to be adapted to the more general news domain and we present evaluation results. We describe our effort to automatically merge the numerous long-form variants referring to the same short-form, while keeping non-related long-forms separate. Finally, we provide extensive statistics on the frequency and the distribution of short-form/long-form pairs across languages.

## 1 Introduction and Motivation

An acronym is an abbreviation formed from the initial letters of the various word elements and read as a single word.[1] Acronyms are formed to speed up and ease communication, mainly to create *words* for concepts frequently used or difficult to describe. Like entities, acronyms have a high reference value, in the sense that they most of the time act as reference anchors of textual content. However, they are not always explicitly defined, which can cause comprehension problems, both for humans and machines. In addition, due to the large number of acronyms – we found over one million when analysing our news data set – the same short-form (SF) can have several conceptually different long-forms (LF) (see Table 1). Even for the same SF-LF pair, many LF variants may exist. In addition to simple wording differences, there can be grammatical inflection forms and cross-lingual variants.

Acronyms are productive words, i.e. new acronyms are created every day, requiring frequent updating of any acronym database. In the first month of applying the tool to our large throughput of multilingual news articles, we identified 66,000 acronyms (before merging variants, i.e. unique SF-LF pairs). After only five months of analysis, the monthly number of newly identified acronym pairs has halved and the number of newly found acronyms seems to be stabilising around this value. We are adding these new acronyms to our multilingual dataset every day and we plan to publicly release the more frequently occurring ones in regular intervals as part of the multilingual name variant resource JRC-Names (Steinberger et al. 2011), which currently predominantly contains person names. This dataset can be used for named entity recognition and other natural language processing tasks, including information retrieval, question answering, summarisation and machine translation.

---

[1] See http://dictionary.reference.com/help/faq/language/t08.html to distinguish acronyms from related concepts such as *initials* and *contractions*.

| |
|---|
| Found in English text<br>    capital adequacy ratio<br>    Capital Adequate Ratio<br>    Capital Adequacy Ration<br>    Capital Adequacy Returns<br>    Center for Autism Research<br>    central African Republic<br>    Certified Automotive Recycler Program<br>    Commission for Aviation Regulation<br>    Confederations of Africa Rugby<br>    Cordilleral Administrative Region<br><br>Found in French text<br>    Caisse Autonome des Retraites<br>    capacité africaine contre les risques<br>    Cellule d'Action Routière<br>    Collectif d'artistes de reggae<br>    Collectivité d'accueil régionale<br>    Comité d'Action pour le Renouveau<br>    Communauté d'agglomération de Rufisque<br><br>Found in German text<br>    Centers for Automotive Research<br>    Central African Republic<br>    chimären Antigenrezeptoren<br>    Computer Assisted Reporting<br><br>Found in Italian text<br>    Cogenerazione ad Alto Rendimento<br>    Computer Assisted Reporting<br>    consumo annuo di riferimento |

**Table 1.** Multilingual examples of acronym long forms for the short form CAR

For acronym recognition, we use the simple and efficient algorithm which was initially developed by Schwartz & Hearst (2003) for the recognition of biomedical abbreviations in English text, but we adapted it for our purposes.

Our contributions are (a) the adaptation of the method to another text type (news); (b) the application to over twenty languages; (c) the generation of highly multilingual statistics on acronym use and on (d) acronym SF ambiguity; and (e) the automatic grouping of LF variant forms.

We first present related work (Section 2), then present our adaptation of the original algorithm, together with recognition statistics and evaluation results (3). We then describe our method to group LF variants (4). We finish by summarising and by pointing to future work (5).

## 2 Related Work

Since the pioneering achievement of Taghva and Gilbreth (1999), a significant amount of work has been completed in the domain of abbreviation processing. Focusing almost exclusively on the bio-medical domain and on the English language, research has developed into three main directions: acronym extraction and mapping to their full forms; acronym variant clustering; and, more recently, acronym disambiguation. We report here on the first two.

With regard to acronym extraction, existing approaches can be divided into four main categories, as suggested by Torii et al. (2007) in their comparative study: *alignment-based* approaches, which exploit the fact that SF and LF show letter or string ordered similarities; *collocation-based* approaches, which exploit the fact that SF and LF frequently occur together and can be considered as collocations; *pattern/rule-based* approaches, which explore regularities of abbreviation conventions; and, finally, *machine-learning* approaches, most of which supervised. Major representatives of these approaches are, respectively: Schwartz and Hearst (2003), whose letter matching algorithm proved to be, despite its simplicity, very efficient; Okazaki and Ananiadou (2006), who address the problem as a term recognition task and perform acronym extraction using statistical co-occurrence evidence in large text collections; Pustejovsky *et al.* (2001), Wren and Garber (2002) and Adar (2004), who look at regular patterns in occurrences of acronyms and manually design templates for their extraction; and Chang *et al.* (2002) and Nadeau and Turney (2005) who apply supervised machine learning algorithms after pre-selection of acronym candidates through the use of Longest Common Subsequence for the former, the use of heuristics for the latter. Although not comparable because focusing on different acronym sub-types (showing different levels of difficulty), these methods perform overall quite well and one can consider the extraction-recognition step a mature technology in the domain of English biomedical literature.

However, not much work exists for languages other than English. Kompara (2010) describes some preliminary work on Slovene, English, French and Italian, while Kokkinakis and Dannélls (2006) investigate the specificity of Swedish – a compounding language – with regard to acronym extraction and present good results obtained thanks to an approach similar to that of Nadeau and Turney (2005). The work showing most similarity with ours is that by Hanh *et al.* (2005). Applying Schwartz and Hearst's algorithm on textual data retrieved from the web in English, German, Portuguese and Spanish, they present a method to align acronyms and their definitions across languages, thanks to an interlingual representation layer. They explore interlingua phenomena and report statistics on the four languages they consider. As opposed to this work, we consider a wider range of languages and we do not intend to use any interlingua.

Finally, it is worth mentioning work on acronym variant clustering: Okazaki *et al.* (2010) present a method to gather similar English acronym expansions based on hierarchical clustering applied over a *pseudo* distance metric. This distance corresponds to a conditional probability, itself computed through binary classification based on various string similarity metric features. Combining all features, they obtain an F-measure of 0.89, noticing that the n-gram similarity was contributing most to the efficiency of the conditional probability. Looking at the same problem, Adar (2004) applies a variant of k-means clustering using the cosine similarity measure over acronym expansion trigrams, and then refined the obtained results taking into account the MeSH category available for each initial n-gram cluster, eventually reaching very good results.

## 3 Multilingual Acronym Extraction

### 3.1 Recognition Algorithm

We use the algorithm presented by Schwartz & Hearst (2003), with minor modifications, mostly consisting of post-processing and filtering the results. In simple words, the algorithm recognises short uppercase expressions between brackets (the SF) and searches in the left-handside con-text for the letters used in the SF. At least the first letter must be word-initial. Unlike Schwartz & Hearst, we do not currently recognise acronym pairs of the format SF (LF) as these are much rarer (in our dataset, less than 10% of all occurrences) and we achieve high recall due to the sheer size of our dataset.

Here are some more details about the algorithm proposed by Ariel & Schwartz: SFs are valid candidates only if they consist of at most two words and if they are between 2 and 10 characters long. If the expression in parentheses is longer, they assume the pattern SF (LF). LF candidates must appear in the same sentence and they must be adjacent to the SF. Regarding their length (the search window), they must not be longer than (a) twice as many words as there are characters in the SF, or (b) the number of characters in the SF plus five words, whichever is the smaller (i.e. $min(|A|+5,|A|*2)$ words, with $|A|$ being the number of characters of the SF).

After applying this pattern to text, we filter the resulting acronym pairs to reduce noise and to avoid unwanted acronym pairs, eliminating cases where either the SF or the LF satisfies any of the following conditions:
a) SFs with currency symbols;
b) SFs with punctuation marks other than hyphens, with quotation marks and word-final apostrophes;
c) SFs starting with a single letter followed by a space;
d) SFs having no uppercase letters.

We additionally eliminate acronyms with LFs satisfying any of the following conditions:
e) LFs excluding white spaces (one-word LFs).

Furthermore, SFs must not:
f) be part of a multilingual stop word list consisting of closed class words (mostly determiners), days of the week or the month and individual words like *north*. Our mixed language stop word list contains about 300 words.

These rules are being applied continuously to large numbers of news texts in the 22 languages of the Europe Media Monitor (EMM) which use the Latin alphabet. EMM processes a current average of 175,000 news articles per day in 70 languages (Steinberger et al. 2009). All acronym pairs are stored, together with meta-information such as date, language, news source and news category, allowing the preparation of detailed statistics.

### 3.2 Multilingual Evaluation

We manually annotated acronyms in 400 articles each in the seven languages Czech, English, French, German, Hungarian, Romanian and Spanish. 200 of these articles were selected randomly (spread over time). The other 200 were selected if our patterns matched at least one acronym pair, to ensure that there is a reasonable number of acronym occurrences to evaluate. The evaluation results in Table 2 show that the performance across languages is rather good and consistent. In comparison, Schwartz & Hearst (2003) report a precision of 0.95 and a Recall of 0.82 when applying their algorithm to the biomedical domain. We conclude that the algorithm works well for a variety of languages, and pre-sumably for all languages using an alphabetic writing system distinguishing lower and upper-case letters.

| ISO | Language | Nº | Prec. | Rec. | F1 |
|---|---|---|---|---|---|
| Cs | Czech | 267 | .96 | .90 | .93 |
| De | German | 274 | .94 | .92 | .93 |
| En | English | 404 | .97 | .91 | .93 |
| Es | Spanish | 339 | .93 | .88 | .90 |
| Fr | French | 371 | .87 | .83 | .85 |
| Hu | Hungarian | 318 | .98 | .96 | .92 |
| Ro | Romanian | 277 | .93 | .91 | .92 |

**Table 2:** Acronym recognition performance results for seven languages (Language ISO code; Number of acronyms evaluated; Precision; Recall; F1 measure).

The major reason for *non-recognition* (lowering Recall) are cases where the acronym's SF is in a different language from the LF, such as in the German *Vereinigte Nationen (UNO)*, where the German LF is followed by the English SF. However, there is a non-negligible number of cases where such cases get coincidentally recognised correctly. Such a lucky case is *Namibische Rundfunkanstalt (NBC)*, where *NBC* stands for the English equivalent *Namibian Broadcasting Corporation*.

The major source of *wrongly recognised* acronym pairs, across all languages, are generic SFs such as the title *CEO* (Chief Executive Officer) or party acronyms such as *PS* (Parti Socialiste) following person names, leading to the erroneous recognition of the acronym pairs like the following: <u>S</u>tephan <u>D</u>orgerloh *(SPD)*; <u>C</u>harl<u>e</u>s <u>O</u>tieno *(CEO)*; <u>c</u>onsists of Pi<u>e</u>ter van <u>O</u>ord *(CEO)*. Some of these cases are hard to avoid. It might therefore be useful to produce lists of such SFs and to filter them additionally, e.g. by combining the recognition patterns with a named entity recognition tool or by training classifiers to get rid of unwanted LFs. It might also be possible to exploit the fact that these SFs occur with unusually high numbers of different LFs, but care must be taken not to also exclude the good LFs. In our evaluation, we came across small numbers of such SFs, leading however to many wrongly recognised acronym pairs.

| ISO | Language | AA distrib. | AS/AA | AA/PU | PO/AA *100 | PO/PU | PU f=1/PU | PU f≥10/PU | PU f≥100/PU | Avg. LF/SF |
|---|---|---|---|---|---|---|---|---|---|---|
| Ca | Catalan | 0.2% | 16.5% | 22 | 21.2 | 4.66 | 61.4% | 6.03% | 0.63% | 1.75 |
| Cs | Czech | 1.5% | 7.5% | 101 | 8.9 | 8.95 | 49.6% | 10.55% | 1.40% | 2.26 |
| Da | Danish | 3.3% | 2.8% | 330 | 3.0 | 9.92 | 55.2% | 13.99% | 1.16% | 1.81 |
| De | German | 12.7% | 10.6% | 69 | 13.3 | 9.09 | 56.4% | 8.33% | 0.99% | 3.52 |
| En | English | 25.1% | 16.4% | 29 | 26.2 | 7.51 | 58.8% | 7.55% | 0.90% | 3.75 |
| Es | Spanish | 11.9% | 21.8% | 38 | 30.7 | 11.64 | 58.2% | 8.92% | 1.26% | 3.31 |
| Et | Estonian | 0.9% | 3.5% | 96 | 3.9 | 3.76 | 62.7% | 5.35% | 0.40% | 2.31 |
| Eu | Basque | 0.0% | 2.6% | 69 | 2.9 | 1.98 | 68.9% | 2.20% | 0.00% | 1.82 |
| Fi | Finnish | 2.2% | 1.3% | 320 | 1.4 | 4.33 | 64.8% | 6.87% | 0.45% | 2.36 |
| Fr | French | 8.8% | 19.3% | 23 | 28.8 | 6.59 | 61.5% | 6.86% | 0.71% | 3.89 |
| Hu | Hungarian | 2.7% | 7.7% | 93 | 9.5 | 8.79 | 56.5% | 8.48% | 1.12% | 2.43 |
| It | Italian | 4.8% | 2.8% | 76 | 3.2 | 2.48 | 71.3% | 3.02% | 0.12% | 2.98 |
| Lt | Lithuanian | 1.0% | 16.5% | 47 | 22.3 | 10.43 | 52.7% | 10.48% | 1.40% | 2.73 |
| Lv | Latvian | 0.9% | 21.4% | 45 | 32.4 | 14.67 | 54.1% | 11.48% | 2.33% | 3.18 |
| Nl | Dutch | 4.2% | 6.9% | 90 | 8.0 | 7.25 | 59.6% | 7.54% | 0.74% | 2.06 |
| No | Norwegian | 1.4% | 5.5% | 87 | 6.2 | 5.41 | 61.2% | 6.90% | 0.64% | 2.03 |
| Pl | Polish | 2.5% | 3.3% | 118 | 3.9 | 4.65 | 55.7% | 7.36% | 0.49% | 2.27 |
| Pt | Portuguese | 4.9% | 20.4% | 47 | 27.9 | 13.13 | 46.3% | 11.08% | 1.52% | 2.91 |
| Ro | Romanian | 5.7% | 10.4% | 60 | 13.5 | 8.11 | 59.2% | 7.86% | 1.15% | 4.32 |
| Sl | Slovene | 1.1% | 7.6% | 67 | 9.2 | 6.16 | 55.3% | 8.91% | 0.80% | 2.62 |
| Sv | Swedish | 4.0% | 2.2% | 289 | 2.4 | 6.96 | 58.7% | 10.60% | 0.85% | 1.89 |
| Sw | Swahili | 0.1% | 13.6% | 26 | 16.6 | 4.32 | 77.2% | 3.21% | 0.52% | 1.94 |
| TOTAL | | 100% | 13.0% | 44 | 18.6 | | 58.7% | 7.85% | 0.95% | 3.40 |

**Table 3:** Statistics on acronym recognition in 22 languages, showing the distribution of articles per language (AA distrib.); the percentage of articles containing at least one acronym (AS/AA); the n° of articles that needs to be parsed to find a new unique acronym (AA/PU); the n° of acronym occurrences per 100 articles (PO/AA*100); the average n° of times a (unique) acronym was reused (PO/PU); the percentage of acronyms that were found only once (PU f=1/PU), at least 10 times (PU f≥10/PU), at least 100 times (PU f≥100/PU); the average number of LFs per SF.

### 3.3 Multilingual Recognition Statistics

We applied the method described in Section 3.1 to many million news articles in 22 languages and produced various types of statistics. These are shown in Table 3. When looking at statistics on, for instance, how many acronyms are used in the different languages, we have to bear in mind that these statistics are biased to some extent by the choices we have made. For instance, we only identify acronym pairs of the form LF (SF), while some languages may more frequently use the inverse order SF (LF) or other alternatives such as *LF, SF* (i.e. the short form is shown inside the text, separated by a comma) or *SF, acronym for LF* (i.e. explicitly mentioning in the text that SF is the acronym for LF). All the numbers in Table 3 refer to successfully recognised acronyms, i.e. after the filtering process described in Section 3.1. When counting unique acronym pairs (PU – pairs unique) or unique SFs, we strictly distinguish case and we consider space and punctuation. For instance, *UNO*, *Uno* and *U.N.O.* are three different SFs. Acronym pair *occurrences* without distinguishing uniqueness are referred to as PO (pairs occurrences). We furthermore use the abbreviations AA for *all articles* analysed and AS for *selected articles*, i.e. only those in which we found acronyms. The highest and the lowest value in each of the columns in Table 3 is written in boldface to give an idea of the range of values.

The first column with numerical contents gives an indication on the relative amount of news text we have analysed. The next column shows that the ratio of news articles AS in which good acronyms (acronyms passing the filtering process) were found, compared to all news articles analysed (AA), is 13%. However, there are enormous differences from one language to the other, with Spanish, Latvian and Portuguese having the highest density of acronyms and Finnish, Swedish and Basque having the lowest.

The third column summarises how many news articles need to be analysed to find a new (i.e. unique) acronym. The fourth column shows how many acronym pair *occurrences* (i.e. non-unique) there are per 100 articles analysed. The fifth column depicts the ratio between unique acronyms PU compared to all acronyms found (PO), thus giving an indication of the number of repetitions of acronyms in the corpus. The sixth column presents the ratio of acronym pairs that have been found exactly once in the corpus (almost 60%), while the next two columns give an indication of how many acronyms have been found at least 10 times or at least 100 times in the corpus. Note that the numbers in Table 3 refer to acronym pairs *before* the merging of acronym variants (described in Section 4). The last column provides the ratio between the number of LFs for the same SF, considering *all* SFs. We thus see that there is an average of 3.4 LFs for each SF. When considering only those SFs that are ambiguous at all (i.e. ignoring SFs that are found with only one LF), the ratio is 6.87.

The statistics on the average number of different SFs for the same unique LF (i.e. the inverse ratio) is less interesting as there are only 1.08 different SFs for the same LF. When considering only the *ambiguous* LFs, the ratio is 2.23, i.e. there are just over two SFs for the same LF. The two different SFs are typically due to varying case, due to plural formation (*ROV* and *ROVs* for *Remotely Operated Vehicles*) or due to punctuation (e.g. *UP* and *U.P.* for *Uttar Pradesh*). However, occasionally, there are also more fundamental differences in the LFs. For instance, in Italian texts, we found the following three acronyms *AUSTRADE*, *Austrade* and *ATC*, all representing the same LF *Australian Trade Commission*.

### 4 Merging related acronym variants

Having identified hundreds of thousands distinct acronym pairs, it is necessary to structure this dataset. We do this by grouping together conceptually related variant LFs belonging to the same SF.

|    |         | N° unique LFs | N° unique SFs | N° LF clusters | N° LF clusters ≥ 2 | Precision | Recogn. Error | Border error |
|----|---------|---------------|---------------|----------------|---------------------|-----------|---------------|--------------|
| DE | German  | 947           | 57            | 402            | 110                 | 0.99      | 0.07          | 0.09         |
| En | English | 955           | 26            | 411            | 144                 | 1.00      | 0.03          | 0.08         |
| Fr | French  | 662           | 21            | 276            | 89                  | 0.99      | 0.05          | 0.03         |
| It | Italian | 576           | 34            | 142            | 55                  | 1.00      | 0.05          | 0.09         |

**Table 4:** Evaluation results for the clustering (separately for each language) of all LFs having the same SF.

### 4.1 Clustering of acronym variants

Given that there are many SFs for which a variety of (relevant and conceptually related) LFs exist, we cluster – separately for each language – all LFs having the same SF. By setting an empirically determined threshold for intra-cluster similarity (or cluster homogeneity), we can group related LFs while keeping unrelated ones separate. We apply binary (hierarchical) group-average clustering. The clustering is based on a pair-wise string similarity for each LF pair in the set. This string similarity is a normalised Levenshtein edit distance where the number of required insertions, deletions and substitutions is divided by the number of characters of the longer LF, yielding a distance value D between 0 and 1. The string similarity S is then the inverse value 1/D. The intra-cluster similarity threshold is set empirically, separately for each language, by optimising it on a development set. For each acronym pair cluster, we choose the most frequently found LF as the representative acronym name.

### 4.2 Evaluation of the clustering

For the evaluation, we manually selected a small number of widely known acronym SFs, for which we could expect that they would be present in each of the languages. Examples are IAEA (International Atomic Energy Agency), IMF (International Monetary Fund), CAR (Central African Republic), ECB (European Central Bank) and FIFA (Fédération Internationale de Football Association), and their respective translations in the four languages (e.g. German EZB and IAEO). This was to make the results comparable across languages. For the rest (the majority), we selected SFs that existed in each of the languages, without knowing whether they would be related across languages and whether the LFs would be similar. This selection was made in preparation of our future work on clustering LF variants *across* languages if they have the same SF.

Table 4 summarises the evaluation results for the acronym LF clustering step for English, French, German and Italian (languages for which we had evaluation volunteers). The first three columns show the number of SF clusters evaluated (unique SF), the number of LFs that had been found and evaluated for these SFs (unique LFs), as well as the number of distinct clusters identified by the clustering algorithm and evaluated (LF clusters). Comparing the third column with the fourth column (clusters ≥ 2) shows that about two thirds of the acronym pairs were not clustered at all and remained single acronyms.

| Agenzia internazionale per l'energia atomica (AIEA) |
|---|
| agenzia delle Nazioni Unite per l'energia atomica |
| Agenzia di controllo sul nucleare delle Nazioni Unite |
| Agenzia internazionale  energia atomica |
| Agenzia Internazionale delEnergia Atomica |
| Agenzia internazionale dell'Onu per l'energia atomica |
| Agenzia internazionale delOnu per energia atomica |
| Agenzia internazionale Energia atomica |
| Agenzia Internazionale Onu per l'Energia Atomica |
| agenzia Internazionale per Energia Atomica |
| Agenzia internazionale per il nucleare |
| Agenzia Internazionale per la Sicurezza Nucleare |
| Agenzia internazionale per l'energia atomica Onu |
| Agenzia nucleare delOnu |
| Agenzia Onu per il Nucleare |
| Agenzia Onu sul nucleare |
| Agenzia per l'Energia atomica |
| Agenzia per l'energia nucleare Onu |
| all'Agenzia internazionale dell'energia atomica |
| all'Organizzazione iraniana dell'energia atomica |
| Atomic Energy Agency |
| Atomica delle Nazioni Unite |
| dell'Agenzia dell'Onu sul nucleare |
| dell'Agenzia delle Nazioni Unite per l'energia atomica |
| dell'Agenzia di controllo sul nucleare delle Nazioni Unite |
| dell'agenzia nazionale per l'energia atomica |

**Table 5.** Subset of LF variants for the Italian SF AIEA, equivalent to English IAEA – International Atomic Energy Agency. All forms were found in real-life news texts.

The precision was evaluated keeping an application-centred approach in mind. Within the framework of ENM, the purpose of the acronym recognition and of the long-form clustering is (a) to display to the users name-like entities as meta-information to news articles and (b) to use these extracted 'entities' as anchors to establish links between related documents (eventually also across languages). For that purpose, we evaluated the precision generously, accepting acronym pairs as rightfully belonging to the same cluster if the intention of the journalist seems to have been to refer to the same entity, even if the acronym LF was not perfectly captured. For that reason, we show recognition error rates separately in Table 4: The column *Recognition Error* describes cases where the system captured non-acronyms or the LFs did not belong to the SF. The column *Border Error* reflects cases where the acronym was detected, but the border of the LF was identified wrongly (e.g. recognising the string *assisted by the International Energy Atomic Agency* for the SF *IAEA*. In such a case, if the erroneous LF was placed in the correct cluster, it was annotated as being correct for clustering, but it was also marked as a *border error*. Journalists are sometimes very lax in their usage of names (see Table 5). It is our intention to capture these references even if the naming may in itself be wrong.

In summary, we find that the clustering process works surprisingly well and that it manages to group LF variants with the same SF, while only rarely excluding LFs that should also be grouped with the cluster. The cases where LFs that refer to the same real-world entity are excluded from a cluster are usually those where the LF differs substantially from those of the entries in the cluster, making it almost impossible to automatically merge the variants. For instance, the German equivalences for *Common Agricultural Policy (CAP)*: *gemeinsame Landwirtschaftspolitik* and *Gemeinsamen Europäischen Agrarpolitik* (GAP) are so different that we do not expect these variants to be recognised automatically without making use of the context of the acronym.

## 5 Conclusion and future work

Acronyms are important referential text elements with high information content that are useful for a whole range of text processing applications. We have shown that an existing English language acronym recognition pattern from the biomedical domain can be adapted successfully to the news domain and to 22 languages from different language families, yielding over one million acronym short-form/long-form pairs. The method works well, for all languages using an alphabetic writing system and distinguishing case. Case is important (a) to select the more promising acronym pairs, thus excluding possible false positives, and also (b) to detect the beginning of the LF string. While we suspect that the method will work well with languages using for instance the Cyrillic or Greek alphabets, it will probably not work well for languages using the Arabic or Hebrew scripts because these do not distinguish case. Clustering turned out to be an efficient method to group acronym spelling variants and separating non-related acronym long-forms coincidentally having the same short-form.

We are interested in categorising the multilingual acronym collection into acronym subtypes such as organisations, programmes (e.g. FP7), stock exchange terminology (e.g. DOW), etc. As our biggest interest are organisation names, we have built a rule-based categoriser using dictionaries with organisation name parts (e.g. *bank, organisation, international, club*, etc.). We believe that, in order to categorise strings in 22 different languages, it is faster to establish and apply such dictionaries than it would be to annotate data in each of the languages and to train a machine learning classifier, but future experiments will show.

The acronym dataset we have created opens up further research avenues. The most interesting challenge probably is how to automatically link acronym long forms across languages. We have several fundamentally different solutions in mind on how to achieve this and we will tackle this task next.

Regarding the recognition of acronyms, it would be interesting to improve the acronym extraction by merging our current method with co-occurrence statistics, which would mostly benefit the recognition of cross-language SF-LF pairs.

Finally, we are interested in recognising and disambiguating acronym SFs that are not accompanied by their LFs, using the local context.

## Acknowledgments

The authors gratefully acknowledge the support of the ERC Starting Grant MultiJEDI No. 259234.

## References

Adar Eytan. 2004. SaRAD: a Simple and Robust Abbreviation Dictionary. BioInformatics 20:527-533.

Chang Jeffrey T., Hinrich Schütze & Russ B. Altman. 2002. Creating an Online Dictionary of Abbreviations from MEDLINE. Journal of the American Medical Informatics Associations 9:262-272.

Hahn Udo, Philipp Daumke, Stefan Schulz & Kornél Markó. 2005. Cross-language mining for acronyms and their completions from the web. Discovery Science, Springer Berlin Heidelberg. 113-123.

Hiroko Ao & Toshihisa Takagi. 2005. ALICE: an algorithm to extract abbreviations from MEDLINE. Journal of the American Medical Informatics Association 12(.5):576-586.

Karipis George. 2005. Cluto: Software for clustering high dimensional datasets. Internet website (last accessed, June 2008), http://glaros.dtc.umn.edu/gkhome/cluto/cluto/overview (2005).

Kompara Mojca. 2010. Automatic recognition of abbreviations and abbreviations' expansions in multilingual electronic texts. Proceedings of CAMLing. 82-91.

Kokkinakis Dimitrios & Dana Dannélls. 2006. Recognizing acronyms and their definitions in Swedish medical texts. Proceedings of the 5th Conference on Language Resources and Evaluation (LREC). Genoa, Italy.

Okazaki Naoaki & Sophia Ananiadou. 2006. Building an abbreviation dictionary using a term recognition approach. Bioinformatics 22(24):3089-3095.

Okazaki Naoaki, Sophia Ananiadou & Jun'ichi Tsujii. 2010. Building a high-quality sense inventory for improved abbreviation disambiguation. Bioinformatics 26(9):1246-1253.

Park Youngja and Roy J. Byrd. 2001. Hybrid text mining for finding abbreviations and their definitions. Proceedings of the 2001 conference on empirical methods in natural language processing, 126-133.

Pustejovsky James, José Castano, Brent Cochran, Maciej Kotecki & Michael Morrell. 2001. Automatic extraction of acronym-meaning pairs from MEDLINE databases. Studies in health technology and informatics 1:371-375.

Nadeau David & Peter Turney. 2005. A supervised learning approach to acronym identification. Proceedings of the Canadian Conference on Artificial Intelligence. 319-329.

Schwartz Ariel S. & Marti A. Hearst. 2003. A simple algorithm for identifying abbreviation definitions in biomedical text. Proceedings of the PAC on Biocomputing, 451-462.

Steinberger Ralf, Bruno Pouliquen & Erik van der Goot (2009). An Introduction to the Europe Media Monitor Family of Applications. In: Fredric Gey, Noriko Kando & Jussi Karlgren (eds.): Information Access in a Multilingual World - Proceedings of the SIGIR 2009 Workshop (SIGIR-CLIR'2009), pp. 1-8. Boston, USA. 23 July 2009.

Steinberger Ralf, Bruno Pouliquen, Mijail Kabadjov & Erik van der Goot (2011). JRC-Names: A freely available, highly multilingual named entity resource. Proceedings of the 8th International Conference Recent Advances in Natural Language Processing (RANLP'2011), pp. 104-110. Hissar, Bulgaria, 12-14 September 2011.Taghva Kazem & Jeff Gilbreth. 1999. Recognizing acronyms and their definitions. International Journal on Document Analysis and Recognition, 1(4):191-198.

Torii Manabu, Zhang-zhi Hu, Min Song, Cathy H Wu & Hongfang Liu. 2007. A comparison study on algorithms of detecting long forms for short forms in biomedical text. BMC Bioinformatics, 8 (suppl. 9):S5.

Wren J. D. & H. R. Garner. 2002. Heuristics for Identification of Acronym-Definition Patterns within Text: Towards an Automated Construction of Comprehensive Acronym-Definition Dictionaries. Methods of Information in Medicine 41(5):426-434.

.